\documentclass{article}    
\usepackage{spconf,amsmath,epsfig}   
\usepackage{hyperref} 

\newcommand{\phileo}{PhilEO} 

\usepackage{graphicx}         
\usepackage{booktabs}   
\usepackage{xcolor} 
\usepackage{multicol}

\title{\phileo\, Bench: Evaluating Geo-Spatial Foundation Models}   
\name{Casper Fibaek, Luke Camilleri, Andreas Luyts, Nikolaos Dionelis, Bertrand Le Saux}      
\address{European Space Agency (ESA), ESRIN, $\Phi$-lab, Italy}  
\begin{document}             
%
\maketitle                                  
\begin{abstract}           
Massive amounts of unlabelled data are captured by Earth Observation (EO) satellites, with the Sentinel-2 constellation generating 1.6TB of data daily. This makes Remote Sensing a data-rich domain well suited to Machine Learning (ML) solutions. However, a bottleneck in applying ML models to EO is the lack of annotated data as annotation is a labour-intensive and costly process. As a result, research in this domain has focused on Self-Supervised Learning and Foundation Model approaches. This paper addresses the need to evaluate different Foundation Models on a fair and uniform benchmark by introducing the \phileo\, Bench, a novel evaluation framework for EO Foundation Models. The framework comprises of a testbed and a novel 400GB Sentinel-2 dataset containing labels for three downstream tasks, building density estimation, road segmentation, and land cover classification. We present experiments using our framework evaluating different Foundation Models, including Prithvi and SatMAE, at multiple $n$-shots and convergence rates.
\end{abstract}         
\section{Introduction}               
\label{sec:intro}   

\textbf{Problem.} Label-efficient learning is challenging as the performance of deep learning models depends highly on the size and the quality of the training data. 
However, for several real-world applications, labelling large datasets is laborious, expensive, and time-consuming. 
This holds for Earth Observation (EO) and remote sensing~\cite{wangSSLinRS2022}, where the data usually require domain expertise: the biologist and the cartographer see different things. Also, the aspect of the Earth changes over time because of nature (seasonality), man-induced changes, and natural hazards, including volcanic eruptions and floods. For a specific geographical region, this can induce a change of label or not.
Data avalanche in EO is another challenge, as many satellites orbit the Earth, such as the Sentinel-2 constellation~\cite{s2_esa}, and produce a lot of data. Labelling them would require massive human interactions and frequent updates. 

For real-world applications \cite{manasSECO2021, Ayush2021geographyaware}, including land cover mapping, training deep learning models with less labeled data is crucial. 
This can be achieved using Self-Supervised Learning (SSL) \cite{wangSSLinRS2022, SSL4EOS12}, that is learning supervised on a pretext task attached to the \textit{structure of the data itself}.




\begin{figure}[tb]     

\begin{minipage}[b]{1.0\linewidth} 
  \centering   
  \centerline{\epsfig{figure=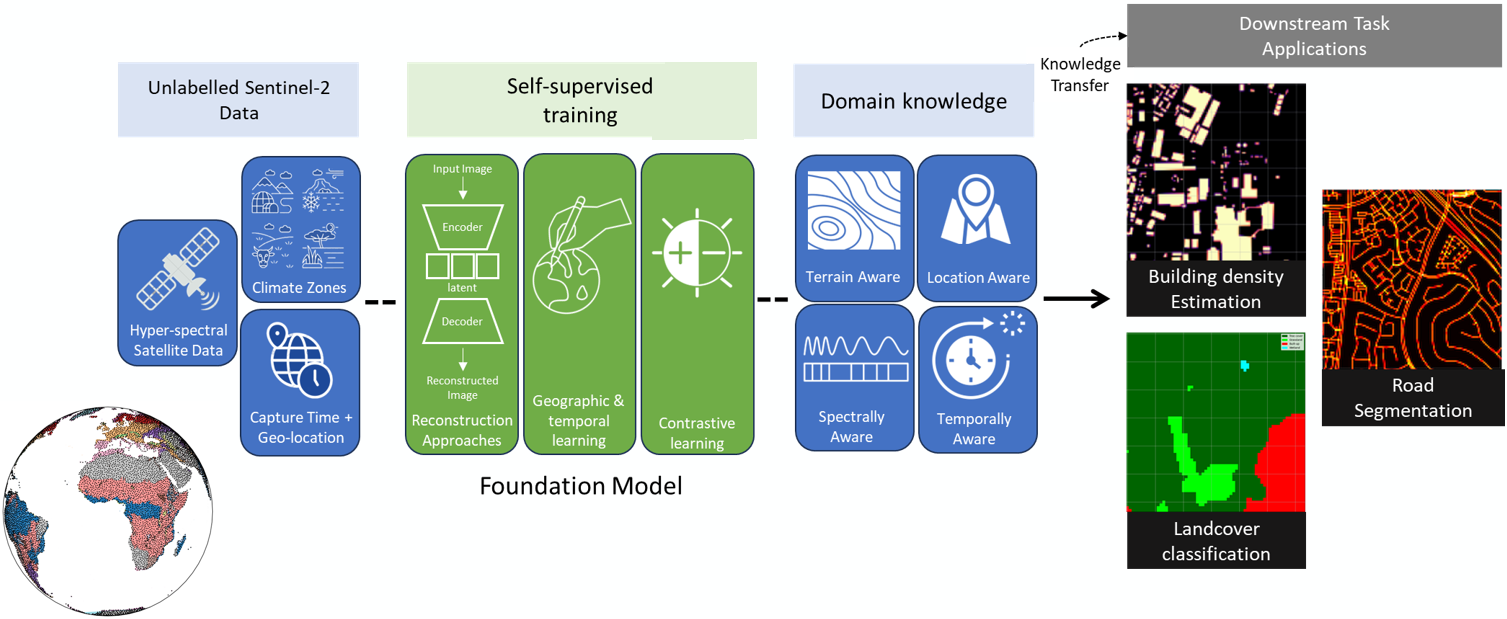,width=8.5cm}}   
\end{minipage}
\caption{The \phileo\, Suite, where complex neural networks are trained on massive data. On the right, the \phileo\, Bench evaluates such foundation models on diverse downstream tasks.}     
\label{fig:Figure1Flowchart}
\end{figure}

\begin{table*}[!tbp]                       
\caption{ \centering Current Foundation Models, trained by self-supervised learning and evaluated on disparate downstream tasks.}    
\vspace{6.6pt}\centering    
\label{tab:OurTable1}      
\vskip -0.0706in 
\vskip -0.0706in
\vskip 0in
\begin{flushleft}              
\begin{scriptsize}   
\begin{sc}         
\begin{tabular}         
{p{0.7cm} p{1.5cm} p{3.cm} p{3.2cm} p{3cm} p{3.9cm}}
\toprule           
\midrule   
\textbf{Name} & \textbf{Dataset} & \textbf{Dataset features} & \textbf{Architecture} & \textbf{Downstream tasks} & \textbf{Benchmark} \\       
\midrule                  
\midrule        
Prithvi & HLS V2 L30 & Coverage: USA, 30m res, 6 bands, Size: 1TB, Time & ViT, MAE, 3D, 100M params, Geolocation & Flood detection, Burn scars, Multi-temp crop & Sen1Floods11: mIoU 88.7\%. HLS Burn Scars: 
IoU 73\% burn class \\          
\midrule           
SatMAE & fMoW & Size: 3.5 TB, Global, 10m res, 8 bands & ViT, MAE, With geolocation, With time & LULC, Multi-label, Building segmentation &  NAIP: Acc: 72\%, BEN: mAP: 82\%, SpaceNet: mIoU: 78\% \\     
\midrule           
SeCo & SeCo & Size: 36 GB, 1M samples, Global, 10m res, RGB & ResNet, MoCo-v2, 23M params, Geolocation & LULC, Change detection, Fine-tuning and LP & BEN: MAP 87\%. EuroSAT: Acc: 93\%. OSCD: F1: 46.9\% \\  
\midrule                
Satlas & SatlasPre- train, 137 classes & Size: 30 TB, Global, 10m res \& 1m, 3 bands (RGB) and 9, With time & Swin Transf, Shifted window self-attention, Labeled data, Geolocation & LULC, Segmentation of roads, buildings, ships. Multi-scale features & UCM: Acc: 99\%, RESISC: 98\%, AID: 88\%, FMoW: F1-score: 44\%, Roads: 87\%, Buildings: 88\% \\    
\midrule           
\midrule     
& & & & \phileo\, (Ours): Building density estimation, Road extraction, LULC & Evaluation framework using annotated Sentinel-2 data that are global, 10m res, 10 bands \\       
\midrule         
\bottomrule       
\end{tabular}    
\end{sc}          
\\               
\vskip 0.05in
HLS = Harmonised Landsat Sentinel-2; LULC = Land Use Land Cover; LP = Linear Probing; BEN = BigEarthNet dataset, OSCD = Onera Satellite Change Detection \\          
\end{scriptsize}               
\end{flushleft}                 
\vskip -0.1in   
\end{table*}  

\textbf{Our contributions.} The main aim of Foundation Models (FM) is to develop effective SSL methods to find good feature representations \cite{SSL4EOS12, AgendaFMs}. Thus, they could be easily adapted to a variety of downstream tasks with good performances, even without much labels in the target domain.
The~\phileo\,~Bench\footnote{Phil in ``\phileo''\, is from the Greek \textit{philia} for love, and EO refers to Earth Observation (EO). Phi is also a wink to ESA $\Phi$-lab ;).}~is~our~proposed evaluation framework for benchmarking geo-spatial FMs on downstream tasks.
The EO Data, the same for all downstream tasks we examine, are global, come from Sentinel-2, and include all bands. The downstream tasks (with corresponding labels) are building density estimation, road segmentation, and land cover classification. 
For the sake of fair comparison, the only variable module is the FM, while the following decoder architecture remains unaltered. The \phileo\, evaluation of a FM is shown in Fig.~\ref{fig:Figure1Flowchart}. In the \phileo\, Bench, we evaluate different EO FMs, including the state-of-the-art Prithvi or SatMAE \cite{nasa2023, cong2023satmae}.

\section{Related work}                
\label{sec:related_work}  

\textbf{Self-supervised learning}  (SSL)~\cite{dosovitskiySSL2016, doerschSSL2017} is a machine learning paradigm that uses unlabeled data to train models to recognize patterns and make decisions that can assist with downstream tasks \cite{wangSSLinRS2022, EmbeddingEarth}. This is done by training the model to solve a task that requires understanding of the data (pretext task) \cite{SSL4EOS12}.
The potential of SSL has been verified also in EO, e.g. Seasonal Contrast (SeCo)~\cite{yuDeepRepRS2018, manasSECO2021}. 
Promising directions for applications have been identified in \cite{wangSSLinRS2022, bergSSLClassif2022, tuia2023artificial} and in \cite{bastani2023satlas, GeospatialFMs}.   
Using SSL has shown to be effective in few-shot learning settings on EO tasks~\cite{manasSECO2021, wangSSLinRS2022}. Thanks to the emergence of large datasets \cite{wangSSLinRS2022, Gl230k}, such as SSL4EO-S12~\cite{SSL4EOS12} which contains 500 GB of S2 data, models have been trained and validated for downstream tasks \cite{sumbulBigEarthNet2021}, for example CORSA~\cite{vitoCORSA2022} or the model USat~\cite{FMsNg}.

\textbf{Foundation Models for Earth science} applications are currently developed \cite{nasa2023}. ClimaX~\cite{nguyen2023climax}, which is for weather and climate, can be tailored to various tasks with different data, coverage, and physical backgrounds. 
SatMAE~\cite{cong2023satmae} is a Vision Transformer (ViT) based on Masked Auto-Encoder (MAE) \cite{maepaper} using temporal and multi-spectral information, trained on Very-High Resolution (VHR) and S2 data at 10m resolution.    
In Table~\ref{tab:OurTable1}, we list characteristics of existing EO FMs.   
Prithvi \cite{nasa2023} is a geospatial FM for EO. 
Trained on US data only, it is not a global/ planetary-scale model and uses six bands at 30m resolution. 
Prithvi and Scale-MAE \cite{scalemae} training is based on SatMAE. 
PRESTO~\cite{tseng2023presto} is a pre-trained Transformer focusing on time-series data. Similarly, CaCo \cite{caco2023} improves upon SeCo \cite{manasSECO2021} by removing  seasonality-invariant training features unsuitable for change detection tasks. 

Geo-Bench \cite{lacoste2023geobench} is a platform for evaluating FMs (with various backbones or training) on downstream tasks.  Different from \cite{lacoste2023geobench} that compiles already existing datasets, the \phileo\, Bench introduces a new global dataset apt to test universality of FMs.

\section{The \phileo\, Evaluation Framework}                      
\label{sec:pagestyle}
The aim of Foundation Models is to improve the performance on downstream tasks \cite{GeographicalKnowledgeDriven, Shi2023LabelEffic, ModelSatCLIP}, while also being generalizable and flexible \cite{NormalizationMatters2023}. However, these models are often evaluated on a range of datasets with different characteristics (size, resolution, locations, satellite sources, and capture dates). There is also a focus on evaluating classification downstream tasks, while omitting image-to-image (such as segmentation) downstream tasks. Therefore, it is challenging to fairly compare the performance of these burgeoning FMs and draw meaningful conclusions. To evaluate FMs, we propose the \phileo\, Bench, an evaluation framework with the aim of providing a flexible, consistent, and fair benchmark for EO Sentinel-2 FMs. 
Our main contributions are: (1) A novel global dataset of S2 images and downstream task labels designed to assess the generalizablity of models across different geographic regions and tasks. (2) A new flexible evaluation framework focused on generating comparable, fair, and reproducible results.

\subsection{\phileo\, Downstream Dataset} \label{sub_sec: datasets}           
\begin{figure}[tb]           
\begin{minipage}[b]{1.0\linewidth}   
  \centering      
  \centerline{\epsfig{figure=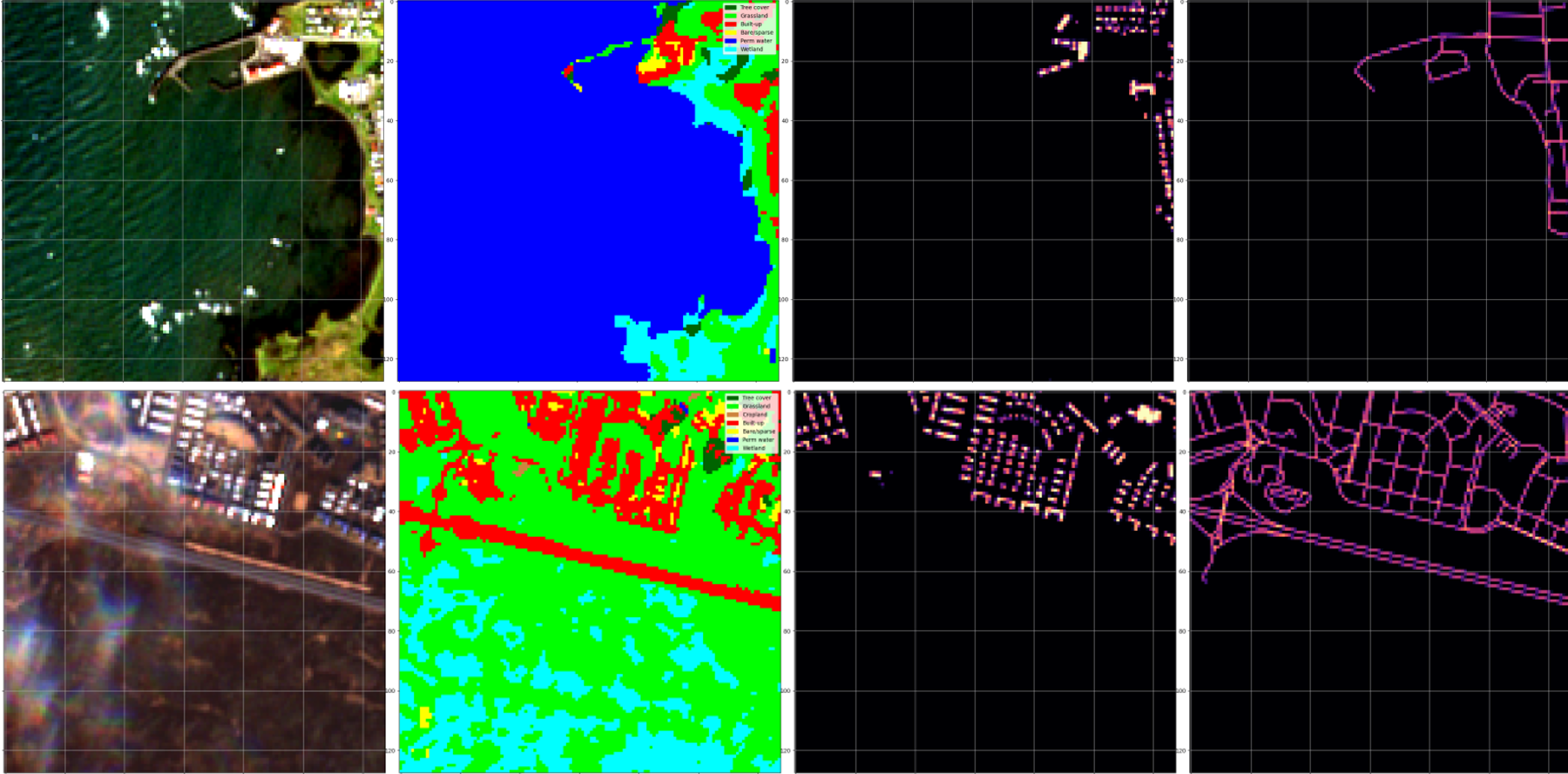,width=6.325cm}}    
\end{minipage} 
\caption{From left: 1) RGB visualisation of S2 patch, 2) Land cover labels, 3) Building Density labels, and 4) Road Density.}      
\label{fig:labels_vis}     
\end{figure}

The \phileo\, dataset is a 400 GB global dataset of S2 images and has labels for roads, buildings, and land cover, where these are the three downstream tasks. The data is sampled from geographically diverse regions around the globe including: Denmark, East Africa, Egypt, Guinea, Europe, Ghana, Israel, Japan, Nigeria, North America, Senegal, South America, Tanzania, and Uganda. Each region has up to 200 tiles of varying sizes. Some locations have been revisited up to $3$ times. The data contain $11$ bands at 10m resolution in the following order: 0-SCL, 1-B02, 2-B03, 3-B04, 4-B08, 5-B05, 6-B06, 7-B07, 8-B8A, 9-B11, and 10-B12 where SCL is the Scene Classification Layer\footnote{For more information about the Sentinel-2 spectral bands, please see: http://docs.sentinel-hub.com/api/latest/data/sentinel-2-l2a}. As shown in Fig.~\ref{fig:labels_vis}, each S2 tile in \phileo\, has a label for each of the downstream tasks:
\begin{itemize}   
    \item \textbf{ROADS}: The labels are expressed as a number of squared meters of roads in a given pixel. The values are between $0$ and $100$, and for a resolution of 10m, this reflect the percentage of coverage. 
    \item \textbf{BUILDINGS}: The labels are expressed as squared meters of buildings. The values are between $0$ and $100$. For 10m resolution, this reflect the coverage (in \%). 
    \item \textbf{LANDCOVER}: Land cover labels are taken from ESA World Cover\footnote{http://worldcover2020.esa.int/data/docs/WorldCover$\_$PUM$\_$V1.1.pdf}: 11 classes, e.g. tree cover and built-up.
\end{itemize}    
All the three downstream tasks can be treated both as segmentation and as classification tasks.

\subsection{Evaluation framework}    \label{sec: eval_framework}   
One of the biggest challenges in evaluating FMs is disentangling the performance impact of various factors such as: model architectures, pre-training tasks, and downstream task training data. 
The effectiveness of a FM can be measured by the quality of its latent representations, and how the key features learnt through the process of pre-training can boost downstream task performance \cite{wangSSLinRS2022, SpGPT}. 
Hence, to provide a \textit{fair} comparison between different FMs within our evaluation framework, we minimize the impact of confounding variables by providing: (1) consistent and repeatable training and evaluation datasets, and (2) a common downstream task head for all the pre-trained models.\begin{itemize} 
    \item \textbf{Dataset creation:} To minimize the impact of variability in the downstream task datasets, a common hold out Test Set for each downstream task was created providing comparable results across all evaluated models. A data partitioning script is also provided that allows for the creation of smaller subsets ($n$-shot or percentage split) from the full \phileo\, Downstream Task training dataset enabling us to evaluate the impact of training set size on model performance. The indexes of the training samples used are automatically saved. Hence, different models can be trained on the identical sub-datasets.
    \item \textbf{One \textit{head} to rule them all:} To minimize the impact of different decoder heads on the downstream task performance, we propose the use of a common decoder head. The latent representations of each of the FMs \cite{nasa2023, Balestriero2023} are fed to a similar decoder. Hence, the comparable performance of a model is a consequence of the effectiveness of the pre-training task and the representational strength of its latent space. For segmentation downstream tasks, we use a multi-convolution decoder based on the U-Net design. For some models, it is required to up-sample or down-sample the features to achieve the desired output image size. For classification downstream tasks, a linear decoder is used \footnote{For ViT models, the cls token is used as the input to the linear decoder.}.
\end{itemize}

Our framework supports two training configurations: (1) \textit{Fine-tuning}, which allows for updating of all downstream task model weights including the FM encoder, and (2) \textit{Linear probing}, where only the decoder head weights are updated, freezing the FM encoder parameters. \phileo\, also contains U-Net \cite{UNetModel}, Mixer \cite{MixerModel}, and ViT \cite{ViTModel} architectures. \phileo\, supports pre-trained models such as Masked Auto-Encoder (MAE) \cite{maepaper} ViT, and Pre-trained U-Nets \cite{UNetModel, KöppenGeigerClimate}, as well as the models Prithvi \cite{nasa2023}, SatMAE \cite{cong2023satmae}, and SeCo \cite{manasSECO2021}. In addition, the testbed should be flexible and easy to use. Hence, an Object Oriented Programming approach is used with an emphasis on modularity, allowing for the easy addition of other downstream tasks, architectures, and pre-trained models.

\section{Evaluating Foundation Models}  
\label{sec:experiments}        
We use the framework of Sec. \ref{sec: eval_framework} to evaluate the performances of our own pre-trained models, including MAE ViTs\footnote{Based on the SatMAE implementation} and Pre-trained U-Nets. 
We also evaluate the state-of-the-art Prithvi, SatMAE, and SeCo. 
In the next sections, we present the experiment, implementation details, and the results.

\subsection{Experiment for the three downstream tasks}      
A $n$-shot transfer learning experiment for the three downstream tasks presented in Sec. \ref{sub_sec: datasets} is set up. 
The experiment aims to examine the impact of different downstream tasks and training set sizes on model performance. 
We also compare the linear probing and fine-tuning transfer learning paradigms.

\subsection{Implementation details for the experiment}        
All models are trained until convergence with an initial learning rate of $0.0001$\footnote{Using a AdamW optimizer and reduce on plateau} and a batch size of $64$. The in-house models we developed are trained on 10-band 10m resolution 128-by-128 patches. The SatMAE models are trained on 10-band 10m resolution 96-by-96 patches. Prithvi is trained on 8-band 30m resolution 224-by-224 patches, while SeCo and ResNet are trained on RGB 10m resolution 224-by-224 patches.

\begin{figure}[tb]   
\begin{minipage}[b]{1.0\linewidth}   
  \centering    
  \centerline{\epsfig{figure=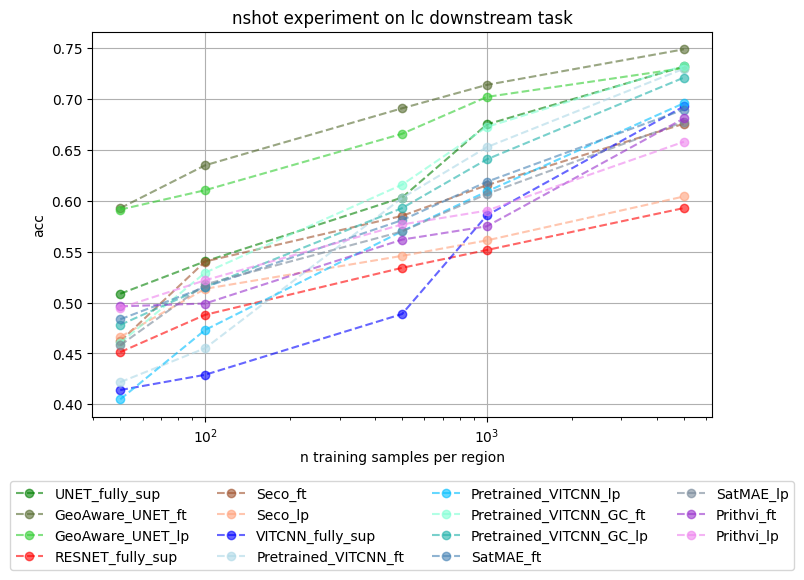,width=8.3cm}}    
\end{minipage}   
\caption{Land cover classification accuracy (acc) using \phileo\, Bench for $n$ samples per region with different architectures and transfer learning paradigms: (1) Linear probing (lp), and (2) Fine-tuning (ft). The \textit{legend} is for both this plot and Fig.~\ref{fig:results_buildings}.}     

\label{fig:results_lc}         
\end{figure}

\subsection{Results and discussion}   
\begin{figure}[tb]       
\vspace{0pt}\begin{minipage}[b] {1.0\linewidth}    
  \centering   
  \centerline{\epsfig{figure=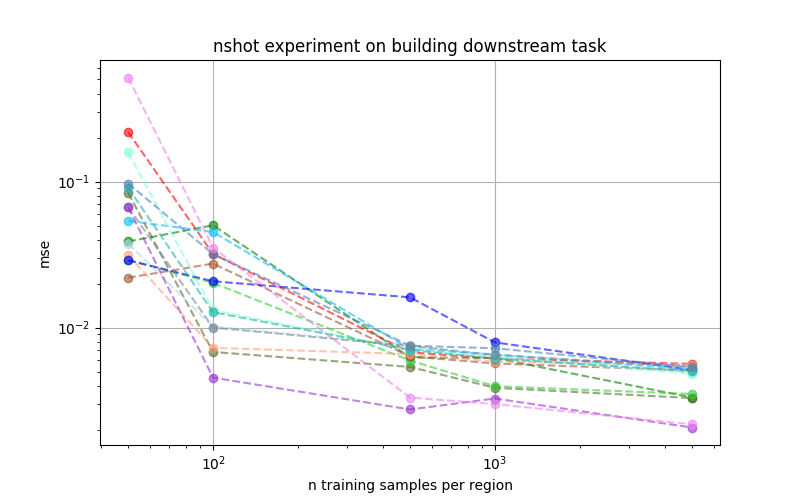,width=8.0cm}}  
\end{minipage}  
\caption{MSE for building density regression using the \phileo\, Bench for $n$ samples per region, evaluating different models.}  
\label{fig:results_buildings}     
\end{figure}

\begin{figure}[tb]     
\begin{minipage}[b]{1.0\linewidth}    
  \centering      
  \centerline{\epsfig{figure=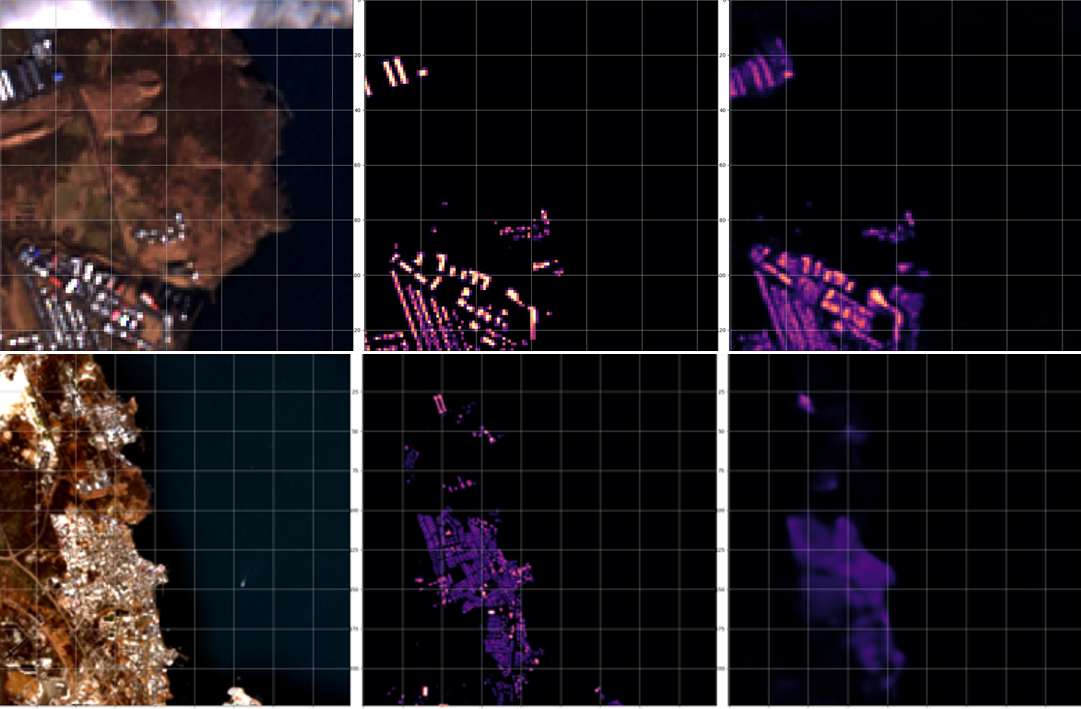,width=3.71cm}}          
\end{minipage}    
\caption{From left: (1) RGB visualisation of Sentinel-2 patch, (2) Building Density labels, and (3) Model predictions. Top row is a U-Net model, while bottom row is the Prithvi model.}     

\label{fig:vis_buldings}            
\end{figure}

From the \phileo~Bench, we pick out two downstream tasks and we report the results we obtain:

\vspace{4pt}  
\textbf{\textit{Image-to-Image Downstream Tasks:}} 
\begin{itemize}    
    \item \textbf{Land Cover}: In Fig.~\ref{fig:results_lc}, we compare the land cover segmentation accuracy of various benchmark and pre-trained models versus the training dataset size. The pre-trained models outperform their randomly initialized fully supervised counterparts, with the pre-trained (fine-tuned) Geo-Aware U-Net consistently giving the best overall performance. The ViTCNN models seem to underperform when compared to the basic U-Net.
    \item \textbf{Building Density}: In Fig.~\ref{fig:results_buildings}, we compare the building density estimation MSE of the benchmark and pre-trained models versus the training dataset size. We observe that the Prithvi-based models result in the lowest MSE scores. However, on visual inspection of the results, it is clear that the outputs generated by the U-Net models are better at picking out \textit{finer} details than the Prithvi models, as shown in Fig.~\ref{fig:vis_buldings}. The discrepancy between the visual results and the MSE scores is because the Prithvi models are trained and evaluated on 30m resolution data, skewing the results in their favour. Fairly evaluating models across different resolutions is a point of future work for the evaluation framework.
\end{itemize}
The results show that state of the art Sentinel-2 FMs can be outperformed by a simple U-Net architecture when applied to image-to-image downstream tasks and $n$-shot transfer learning experiments\footnote{Our code can be found in the GitHub repository \url{http://github.com/ESA-PhiLab/PhilEO-Bench}}. 
These results are not all that surprising when one considers the encoder-decoder structure of these pre-trained models. 
The amount of information that can be stored in the latent space (bottleneck) is limited. 
Therefore, only salient high-level features are compressed into the latent space. 
The lack of low-level information in the latent space makes it challenging to reconstruct fine-grained details in the output image, such as building locations, resulting in poor image-to-image downstream task performance. 
The U-Net skip connections address this by bypassing the bottleneck and providing low-level features to the decoder, allowing for the reconstruction of fine-grained details in the output image.
\par 





\section{Conclusion}    
\label{sec:page}      

We have proposed the \phileo\, evaluation framework which comprises both a new testbed for benchmarking Sentinel-2 FMs and a novel global dataset. 
With this evaluation framework, we hope to promote the advancement of Sentinel-2 FMs by providing a general, fair, and easy to use evaluation tool. 
The experiments we carried out on the \phileo\, Bench have exposed a weakness in the current state of the art EO FMs. By mainly focusing on classification tasks, most existing models are based on the Masked Auto-Encoder (MAE) Vision Transformer (ViT) architecture. 
However, this model architecture does not lend itself well to image-to-image downstream tasks, with even a basic U-Net outperforming models such as SatMAE and Prithvi. 
We believe that this case study underlines the utility of our evaluation testbed and the importance of having a fair comparison between different models.

\section{Acknowledgement}           
We thank for fruitful discussions Leonardo Bagaglini, Giacomo D. Cascarano and Giorgio Pasquali (Leonardo Labs).




\bibliographystyle{IEEEbib}   
\bibliography{refs}

\end{document}